\def\year{2017}\relax
\documentclass[letterpaper]{article}
\usepackage{aaai17}
\usepackage{times}
\usepackage{helvet}
\usepackage{courier}
\usepackage{graphicx}
\usepackage{tabularx}
\usepackage{array}
\usepackage{multirow}
\usepackage{float}
\usepackage{mathrsfs,amsmath}
\usepackage{booktabs}
\newcolumntype{L}[1]{>{\raggedright\let\newline\\\arraybackslash\hspace{0pt}}m{#1}}
\newcolumntype{C}[1]{>{\centering\let\newline\\\arraybackslash\hspace{0pt}}m{#1}}
\newcolumntype{R}[1]{>{\raggedleft\let\newline\\\arraybackslash\hspace{0pt}}m{#1}}
\DeclareMathOperator*{\argmaxA}{arg\,max} 

\begin{document}

\title{Condensed Memory Networks for Clinical Diagnostic Inferencing}

\author{Aaditya Prakash$^{*}$\\ Brandeis University, MA \\ aprakash@brandeis.edu \\ 
\And Siyuan Zhao \\ Worcester Polytechnic Institute, MA \\ szhao@wpi.edu \\
\AND Sadid A. Hasan, Vivek Datla, Kathy Lee, Ashequl Qadir, Joey Liu, Oladimeji Farri\\ $^{*}$Artificial Intelligence Laboratory, Philips Research North America, Cambridge, MA\\ \{firstname.lastname,kathy.lee\_1,dimeji.farri\}@philips.com}

\maketitle

\begin{abstract}
    Diagnosis of a clinical condition is a challenging task, which often requires significant medical investigation. Previous work related to diagnostic inferencing problems mostly consider multivariate observational data (e.g.\ physiological signals, lab tests etc.). In contrast, we explore the problem using free-text medical notes recorded in an electronic health record (EHR). Complex tasks like these can benefit from structured knowledge bases, but those are not scalable. We instead exploit raw text from Wikipedia as a knowledge source. Memory networks have been demonstrated to be effective in tasks which require comprehension of free-form text. They use the final iteration of the learned representation to predict probable classes. We introduce condensed memory neural networks (C-MemNNs), a novel model with iterative condensation of memory representations that preserves the hierarchy of features in the memory. Experiments on the \texttt{MIMIC-III} dataset show that the proposed model outperforms other variants of memory networks to predict the most probable diagnoses given a complex clinical scenario. 

\end{abstract}

\section{Introduction}
     Clinicians perform complex cognitive processes to infer the probable diagnosis after observing several variables such as the patient's past medical history, current condition, and various clinical measurements. The cognitive burden of dealing with complex patient situations could be reduced by having an automated assistant provide suggestions to physicians of the most probable diagnostic options for optimal clinical decision-making.

    Some work has been done in building Artificial Intelligence (AI) systems that can support clinical decision making~\cite{DBLP:journals/corr/LiptonKEW15,choi2015doctor,Choi2016RETAINIP}. These works have primarily focused on the use of various biosignals as features. EHRs typically store such structured clinical data (e.g.\ physiological signals, vital signs, lab tests etc.) about the patients' clinical encounters in addition to unstructured textual notes that contain a complete picture of the associated clinical events. Structured clinical data generally contain raw signals without much interpretation, whereas unstructured free-text clinical notes contain detailed description of the overall clinical scenario. In this paper, we explore the discriminatory capability of the unstructured free-text clinical notes to correctly infer the most probable diagnoses from a complex clinical scenario. In the Clinical Decision Support track\footnote{http://www.trec-cds.org/} of the recent Text REtrieval Conference (TREC) series, clinical diagnostic inferencing from unstructured free texts has been shown to play a significant role in improving the accuracy of relevant biomedical article retrieval~\cite{sadid14,sadid15,sadid16}. We also explore the use of an external knowledge source like Wikipedia from which the model can extract relevant information, such as signs and symptoms for various diseases. Our goal is to combine such an external clinical knowledge source with the free-text clinical notes and use the learning capability of memory networks to correctly infer the most probable diagnosis.
    
    Memory Networks (MemNNs)~\cite{NIPS2015_end_to_end,DBLP:journals/corr/WestonCB14} are a class of models which contain an external memory and a controller to read from and write to the memory. Memory Networks read a given input source and a knowledge source several times (hops) while updating an internal memory state. The memory state is the representation of relevant information from the knowledge base optimized to solve the given task. This allows the network to remember useful features. The notion of neural networks with memory was introduced to solve AI tasks that require complex reasoning and inferencing. These models have been successfully applied in the Question Answering domain on datasets like \emph{bAbi}~\cite{weston2015towards}, \emph{MovieQA}~\cite{tapaswi2015movieqa}, and \emph{WikiQA}~\cite{NIPS2015_end_to_end,DBLP:journals/corr/MillerFDKBW16}. Memory networks are harder to train than traditional networks and they do not scale easily to a large memory. End-to-End Memory Networks~\cite{NIPS2015_end_to_end} and Key-Value Memory Networks (KV-MemNNs)~\cite{DBLP:journals/corr/MillerFDKBW16} try to solve these problems by training multiple hops over memory and compartmentalizing memory slots into hashes, respectively. 
    
    When the memory is large, hashing can be used to selectively retrieve only relevant information from the knowledge base, however not much work has been done to improve the information content of the memory state. If the network were trained for factoid question answering, the memory state might be trained to represent relevant facts and relations from the underlying domain. However, for real world tasks, a large amount of memory is required to achieve state-of-the-art results. In this paper, we introduce Condensed Memory Networks (C-MemNNs), an approach to efficiently store condensed representations in memory, thereby maximizing the utility of limited memory slots. We show that a condensed form of memory state which contains some information from earlier hops learns efficient representation. We take inspiration from human memory for this model. Humans can learn new information and yet remember even very old memories as abstractions. We also experiment with a simpler form of knowledge retention from previous hops by taking a weighted \textbf{average} of memory states from all the hops (A-MemNN). Even this simpler alternative which does not add any extra parameter is able to outperform standard memory networks. Empirical results on the \texttt{\texttt{MIMIC-III}} dataset reveal that C-MemNN improves the accuracy of clinical diagnostic inferencing over other classes of memory networks. To the best of our knowledge, this is the first empirical study to classify diagnosis from EHR free-text clinical notes using memory networks.

\section{Related Work}
    \subsection{Memory Networks}
    Memory Networks (MemNNs)~\cite{DBLP:journals/corr/WestonCB14} and Neural Turing Machines (NTMs)~\cite{graves2014neural} are the two classes of neural network models with an external memory component. MemNN stores all information (e.g.\ knowledge base, background context) into the external memory, assigns a relevance probability to each memory slot using content-based addressing schemes, and reads contents from each memory slot by taking their weighted sum. End-to-End Memory Networks introduced multi-hop training~\cite{NIPS2015_end_to_end} and do not require strong supervision unlike MemNN\@. Key-Value Memory Networks~\cite{DBLP:journals/corr/MillerFDKBW16} have a key-value paired memory and are built upon MemNN\@. The key-value paired structure is a generalized way of storing content in the memory. The contents in the key-memory are used to calculate the relevance probabilities whereas the contents in the value-memory are read into the model to help make the final prediction.
    
    In contrast to MemNN, which uses a content-based mechanism to access the external memory, the NTM controller uses both content- and location-based mechanisms. The fundamental difference between these models is that MemNN does not have a mechanism for the content to be changed in the memory, while NTM can modify the content in each episode. This makes MemNN easier to train compared to NTM.

    Recently, there has been an attempt to incorporate longer contextual memory into the basic Recurrent Neural Networks (RNNs) framework. Stack-Augmented RNN~\cite{joulin2015inferring} proposes interconnecting RNN modules using a push-down stack in order to learn long-term dependencies. They are able to reproduce complicated sequence patterns.~\citeauthor{chung2016hierarchical} (2016) explore multi-scale RNN, which is able to learn a latent hierarchical structure by using temporal representation at different time-scales. These methods are well-suited for learning long-term temporal dependencies, but do not scale well to large memory. Hierarchical Memory Networks~\cite{chandar2016hierarchical} study the use of maximum inner product search (MIPS) to store memory slots in a hierarchy. Our aim is to improve the efficiency and knowledge density of the standard memory slots by exploring a hierarchy of internal memory representations over multiple hops.
    
    Another related class of models are the attention-based neural networks. These models are trained to learn an attention mechanism so that they can focus on the important information on a given input. Applying an attention mechanism on the machine reading comprehension task~\cite{Hermann2015TeachingMT,Dhingra2016GatedAttentionRF,Cui2016AttentionoverAttentionNN,Sordoni2016IterativeAN} has shown promising results. In tasks where inferencing is governed by the input source e.g.\ sentence-level machine translation~\cite{bahdanau2014neural}, image caption generation~\cite{xu2015show}, and visual question answering~\cite{lu2016hierarchical}, the use of attention-based models has proven to be very effective. As attention is learned by the iterative finding of the highly-activated input regions, this is not feasible for a large-scale external memory; however, more research is required in order to achieve attention over memory.

    \subsection{Neural Networks for Clinical Diagnosis}
    The application of neural networks to medical tasks dates back more than twenty years~\cite{baxt1990use}. The recent success of deep learning has drawn broader interest in building AI systems to support clinical decision making.~\citeauthor{DBLP:journals/corr/LiptonKEW15} (2015) train Long Short-Term Memory Networks (LSTMs) to classify 128 diagnoses from 13 frequently but irregularly sampled clinical measurements extracted from the EHR\@. DoctorAI~\cite{choi2015doctor} and RETAIN~\cite{Choi2016RETAINIP} also use time series data for diagnosis classification. Similar to these works, we formulate the problem as multi-label classification, since each medical note might be associated with multiple diagnoses. However, there are two important differences between our work and the previous work. We only consider discharge notes for our experiments, which are unstructured free-texts and do not contain time series data, while they rely on time series datasets where each time series has a fixed number of clinical measurements. Moreover, we train our model in an end-to-end fashion and do not extract any hand-engineered features from the notes, while they resample all time series data to an hourly rate and fill in the gaps created by window-based resampling in clinical measurements. We propose the use of memory networks instead of LSTMs to classify the  diagnoses since the memory component provides the flexibility to learn from an external knowledge source. 

\section{Dataset}
    \texttt{\texttt{MIMIC-III}} (Multiparameter Intelligent Monitoring in Intensive Care)~\cite{johnson2016mimic} is a large freely-available clinical database. It contains physiological signals and various measurements captured from patient monitors, and comprehensive clinical data obtained from hospital medical information systems for over $58K$ Intensive Care Unit (ICU) patients.  We use the \emph{noteevents} table from \texttt{\texttt{MIMIC-III}}: v1.3, which contains the unstructured free-text clinical notes for patients. We use `discharge summaries', instead of `admission notes', as former contains actual ground truth and free-text. Since discharge summaries are written after diagnosisdecision, we sanitize the notes by removing any mention of class-labels in the text.

\begin{table}
    \small
    \centering
\label{mimic-ii}
    \def\arraystretch{1.5}
    \begin{tabularx}{8.5cm}{X}
    \midrule
    \textbf{Medical Note} (partially shown) \\
    \midrule
    Date of Birth:   {\tiny [**2606--2--28**] }     Sex:  M \newline
    Service:  Medicine \newline
    \textbf{Chief Complaint:} \newline
    Admitted from rehabilitation for
    hypotension (systolic blood pressure to the 70s) and
    decreased urine output.
    \textbf{History of present illness:} \newline
    The patient is a 76-year-old male who had
    been hospitalized at the {\tiny[**Hospital1 3007**]} from {\tiny[**8--29**]} through {\tiny[**9--6**]} of 2002
    after undergoing a left femoral-AT bypass graft and was
    subsequently discharged to a rehabilitation facility. \newline
    On {\tiny[**2682--9--7**]}, he presented again to the {\tiny[**Hospital1 3087**]}
    after being found to have a systolic
    blood pressure in the 70s and no urine output for 17 hours.\\
    
    \midrule
    \textbf{Diagnosis} \\
    \midrule
    Cardiorespiratory arrest. \newline
    Non-Q-wave myocardial infarction. \newline
    Acute renal failure. \\
    \midrule
    \end{tabularx}
    \caption{An example \texttt{\texttt{MIMIC-III}} note.}
\end{table}

\begin{figure}
    \includegraphics[width=\linewidth]{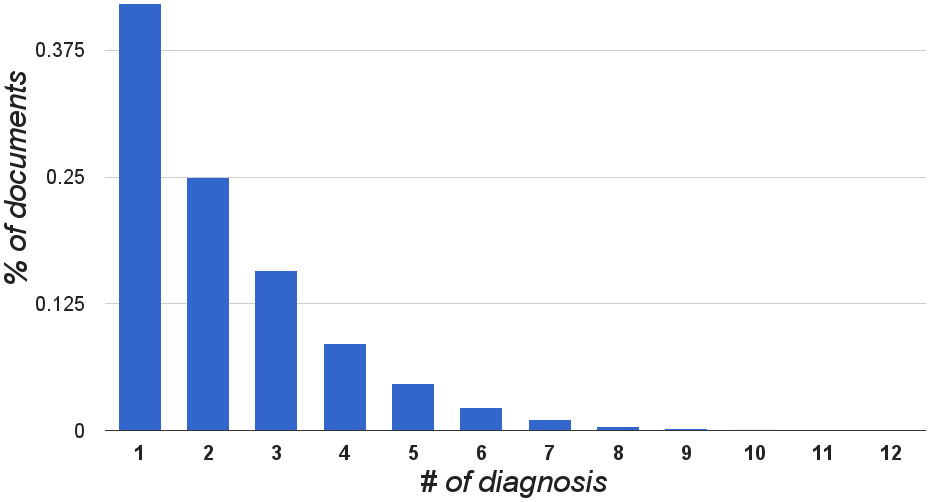}
  \caption{Distribution of number of diagnosis in a note.}
\label{number_of_diagnosis}
\end{figure}

\begin{figure*}
 \centering 
 \scalebox{.9}{\includegraphics[width=\textwidth,scale=0.2]{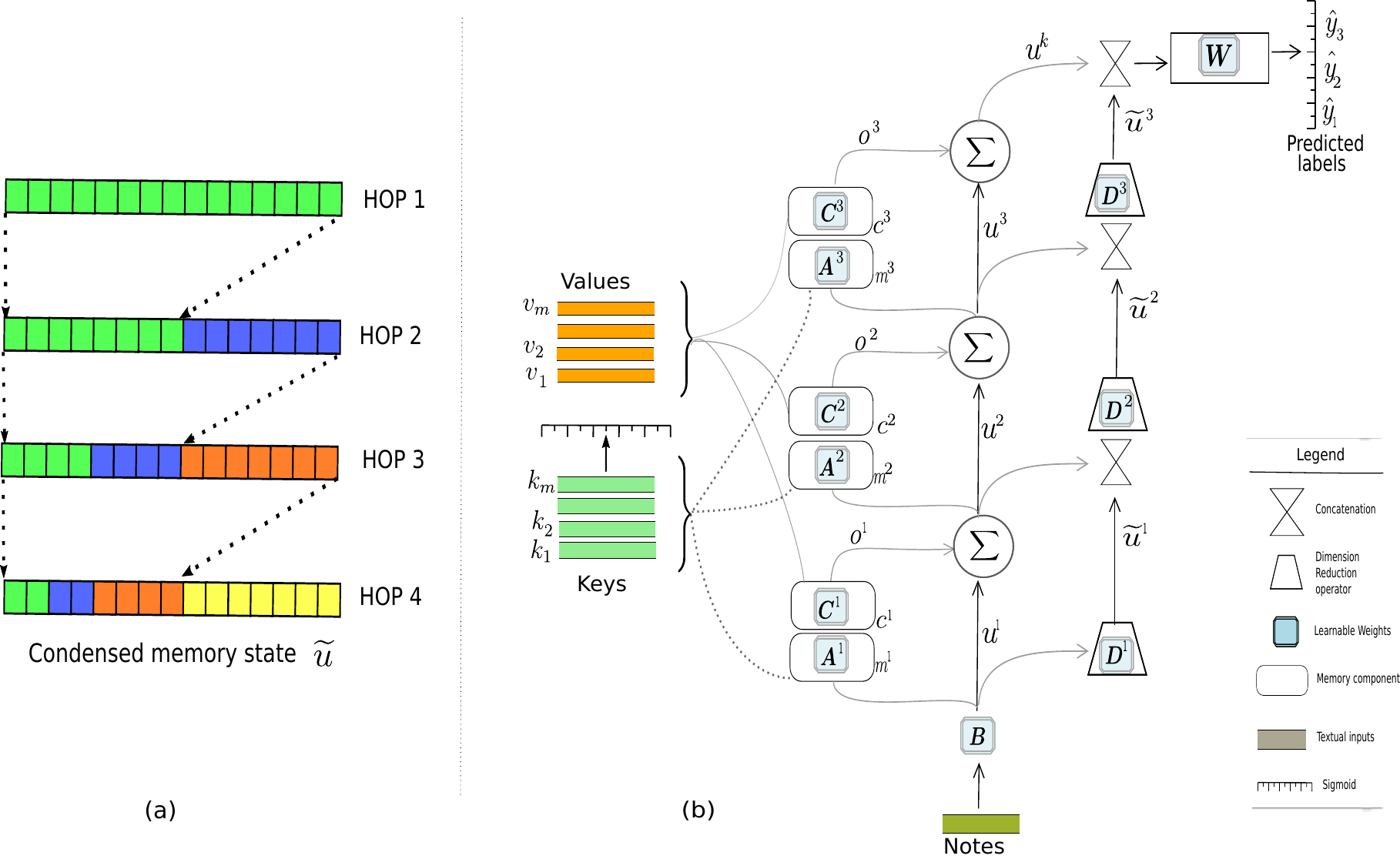} }
  \caption{\small (a) Abstract view of transformation of memory representation over multiple hops. 
  (b) Structural overview of end-to-end model for condensed memory networks.}
\end{figure*}

As shown in Table 1, medical notes contain several details about the patient but the sections are not uniform. We do not separate the sections other than the \emph{DIAGNOSIS}, which is our label. There are multiple labels (diagnoses) for a given note, and a note can belong to multiple classes of diagnoses, thus we formulate our task as a multiclass-multilabel classification problem. The number of diagnoses per note is also not consistent and shows a long tail (Figure~\ref{number_of_diagnosis}). We have taken measures to counteract these issues, which are discussed in the \emph{Memory addressing} section. 

Some diagnoses are less frequent in the data set. Without enough training instances, a model is not able to learn to recognize these diagnoses. Therefore, we experiment with a varying number of labels in this work (see details in the \emph{Experiments} section). 

\subsection{Knowledge Base}
We use Wikipedia pages (see Table 2) corresponding to the diagnoses in the \texttt{MIMIC-III} notes as our external knowledge source. WikiProject Medicine is dedicated to improving the quality of medical articles on Wikipedia and the information presented in these pages are generally shown to be reliable~\cite{trevena2011wikiproject}. 
Since some diagnosis terms from \texttt{MIMIC-III} don't always match a Wikipedia title, we use the Wikipedia API with the diagnoses as the search terms to find the most relevant Wikipedia pages. In most cases we find an exact match while in the rest we pick the most relevant page. We use the first paragraph and the paragraphs corresponding to the \emph{Signs and symptoms} sections for our experiments. In cases where such a section is not available, we use the second and third paragraphs of the page. This happens for the obscure diseases, which have a limited content.

\begin{table}
    \small
    \centering
\label{wiki}
    \def\arraystretch{1.5}
    \begin{tabularx}{8.5cm}{X}
    \midrule
    \textbf{Cardiac arrest} \\
    \midrule
    Cardiac arrest is a sudden stop in effective blood circulation due to the failure of the heart to contract effectively or at all[1]. A cardiac arrest is different from (but may be caused by) a myocardial infarction (also known as a heart attack), where blood flow to the muscle of the heart is impaired such that part or all of the heart tissue dies\ldots \newline
    \textbf{Signs and symptoms}\newline
    Cardiac arrest is sometimes preceded by certain symptoms such as fainting, fatigue, blackouts, dizziness, chest pain, shortness of breath, weakness, and vomiting. The arrest may also occur with no warning \ldots \\
    \midrule
    \end{tabularx}
    \caption{\small Partially shown example of a relevant Wikipedia page.}
\end{table}

\section{Condensed Memory Networks}
The basic structure of our model is inspired by MemNN\@. Our model tries to learn memory representation from a given knowledge base. Memory is organized as some number of slots $m_1, \ldots m_t$. For the given input text i.e.\ medical notes $x_1, \ldots, x_n$, the external knowledge base (wiki pages, wiki titles) $(k_1, v_1), (k_2, v_2), \ldots , (k_m, v_m)$, and the diagnoses of those notes $y$, we aim to learn a model $\mathcal{F}$ such that
    \begin{equation}
        \mathcal{F}(x_n, (k_m, v_m)) = \hat{y} \rightarrow y
    \end{equation}

    We break down this function $\mathcal{F}$, in four parts \textit{I, G, O, R} which are the standard components of Memory Networks.
    \begin{itemize}
        \item \textbf{I: Input memory representation} is the transformation of the input $x$ to some internal representation $u$ using learned weights $\boldsymbol{B}$. This is the internal state of the model and is similar to the hidden state of RNN-based models. In this paper, we propose the addition of a condensed memory state $\widetilde{u}$, which is obtained via the iterative concatenation of successively lower dimensional representations of the input memory state $u$. 
        \item \textbf{G: Generalization} is the process of updating the memory. MemNN updates all slots in the memory, but this is not feasible when the size of the knowledge source is very large. Therefore,  we organize the memory as key-value pairs as described in~\citeauthor{DBLP:journals/corr/MillerFDKBW16} (2016). We use hashing to retrieve a small portion of keys for the given input.  

        \item \textbf{O: Output memory representation} is the transformation of the knowledge $(k,v)$ to some internal representation $m$ and $c$. While End-to-End MemNN uses an embedding matrix to convert memories to learned feature space, our model uses a two-step process because we represent wiki-titles and wikipages as different learned spaces. We learn matrix $\boldsymbol{A}$ to transform wikipages (keys) and $\boldsymbol{C}$ to transform wiki-titles (values). Our choice of wikipages as keys and wiki-titles as values is deliberate -- the input ``medical notes'' more closely match the text of the wikipages and the diagnoses more closely match the wiki-titles. This allows for a better mapping of features; our empirical results validate this idea.

        Let $k$ represents the hop number. The output memory representation is obtained by:
            \begin{equation}
                o^k = \sum_i \text{Addressing}(u^k, m_i^k) \cdot c_i^k  
            \end{equation}
            
        where \textit{Addressing} is a function which takes the given input memory state $u$ and provides the relevant memory representation $m$.

    \item \textbf{R: Response} combines the internal state $u$, internal condensed state $\widetilde{u}$ and the output representation $o$ to provide the predicted label $\hat{y}$. We sum $u$ and $o$ and then take the dot product with another learned matrix $\boldsymbol{W}$. We then concatenate this value with condensed memory state $\widetilde{u}$. This value is then passed through sigmoid to obtain the likelihood of each class. We use sigmoid instead of softmax in order to obtain multiple predicted labels, $\hat{y_1}, \ldots \hat{y_r}$ among possible $R$ labels. Both the memory states $u$ and $\widetilde{u}$ are computed as:
            
            \begin{equation}
                u^{k+1} = u^k + o^k 
            \end{equation}
    
            \begin{equation}
                \widetilde{u}^{k+1}    = u^{k+1} \oplus \boldsymbol{D_1} \cdot \widetilde{u}^k
            \end{equation}
    
            where $\oplus$ denotes concatenation of vectors. 
            
            Our major contribution to memory networks is the use of condensed memory state $\widetilde{u}$ in combination with input memory state $u$ to do the inference. As shown in Figure 2(a), $\widetilde{u}$ is transformed to include the information of previous hops, but in lower dimensional feature space. This leads to a longer term memory representation, which is better able to represent hierarchy in memory. The class prediction is obtained using:

            \begin{equation}
                \hat{y_r} = \argmaxA_{r \epsilon R} {\frac {1}{1+e^{-1 * (\widetilde{u}^{k+1} \cdot \boldsymbol{W})}}} \label{eq:probs}
            \end{equation}

     \end{itemize}
    
     \subsection{Network Overview}
            
            Figure 2(b) shows the overview of the C-MemNN structure. The input $x$ is converted to internal state $u^1$ using the transformation matrix $\boldsymbol{B}$. This is combined with memory key slots $m^1$ using matrix $\boldsymbol{A}$. Memory addressing is used to retrieve the corresponding memory value $c^1$. This value is transformed using matrix $\boldsymbol{C}$ to output memory representation $o^1$. In parallel, memory state $u$ is condensed to half of its original dimension using the transformation matrix $\boldsymbol{D}$. If $u$ is of size $1 \times K$ then $\boldsymbol{D}$ is of size $K \times \frac{K}{2}$. We call this reduced representation of $u$ the condensed memory state, $\widetilde{u}$. This is the end of the first hop. This process is then repeated for a desired number of hops. After each hop, the condensed memory state $\widetilde{u}$ becomes the concatenation of its previous state and its current state, each reduced to half of its original dimension.

\subsection{Averaged Memory Networks}

In C-MemNN, the transformation of $\widetilde{u}$ at every hop adds more parameters to the model, which is not always desirable. Thus, we also present a simpler alternative model, which we call A-MemNN, to capture hierarchy in memory representation without adding any learned parameters. In this alternative model, we compute the weighted average of $\widetilde{u}$ across multiple hops. Instead of concatenating previous $\widetilde{u}$ values, we simply maintain an exponential moving average from different hops: 

    \begin{equation}
         \widetilde{u}^{k+1} =  \widetilde{u}^{k} +  \frac{\widetilde{u}^{k-1}}{2}  + \frac{\widetilde{u}^{k-2}}{4} + \ldots
    \end{equation}
    where, the starting condensed memory state is same as the input memory state $\widetilde{u}^1 = u^1$.

    \subsection{Memory Addressing}

    Key-Value addressing as described in KV-MemNN uses softmax on the product of question embeddings and retrieved keys to learn a relevance probability distribution over memory slots. The representation obtained is then the sum of the output memory representation $o$, weighted by those probability values. KV-MemNN was designed to pick the single most relevant answer given a set of candidate answers. The use of softmax significantly decreases the estimated relevance of all but the most probable memory slot.  This presents a problem for multi-label classification in which several memory slots may be relevant for different target labels.  We experimented with changing softmax to sigmoid to alleviate this problem, but this was not sufficient to allow the incorporation of the condensed form of the internal state $u$ arising from earlier hops. Thus, we explore a novel alternate addressing scheme, which we call gated addressing.  This addressing method uses a multi-layer feed-forward neural network (FNN) with a sigmoid output layer to determine the appropriate weights for each memory slot.  The network calculates a weight value between 0 and 1 for each memory slot, and a weighted sum of memory slots is obtained as before.

    \subsection{Document Representation}
    There are a variety of models to represent knowledge in key-value memories, and the choice of a model can have an impact on the overall performance. We use a simple bag-of-words (BoW) model which transforms each word $w_{ij}$ in the document $d_i = {w_{i1}, w_{i2}, w_{i3}, \ldots, w_{in}}$ to embeddings, and sums these together to obtain the vectors $\Phi(d_i) = \sum_jAw_{ij}$, with $A$ being the embedding matrix. Medical notes, memory keys and memory values are all represented in this way.

\section{Experiments}
    The distribution of diagnoses in our training data has a very long tail. There are 4,186 unique diagnoses in \texttt{MIMIC-III} discharge notes. However, many diagnoses (labels) occur in only a single note. This is not sufficient to efficiently train on these labels. The 50 most-common labels cover 97\%  of the notes and the 100 most-common labels cover 99.97\%. Thus, we frame this task as multi-label classification for top-N labels. We present experiments for both the 50 most-common and 100-most common labels. For all experiments, we truncate both notes and wiki-pages to 600 words. We reduce the trained vocabulary to 20K after removing common stop-words. We use a common dimension of 500 for all embedding matrices. We use a memory slot of dimension 300. A smaller embedding of dimension 32 is used to represent the wiki-titles.
    
    We present experiments for end-to-end memory networks~\cite{NIPS2015_end_to_end}, Key-Value Memory Networks (KV-MemNNs)~\cite{DBLP:journals/corr/MillerFDKBW16} and our models, Condensed Memory Networks (C-MemNN) and Averaged Memory Networks (A-MemNN). We separately train models for three, four and five hops. The strength of our model is the ability to make effective use of several memory hops, and so we do not present results for one or two hops. We did not consider standard text classification models like CNN, LSTM or Fasttext~\cite{joulin2016bag} as they do not allow incorporating external memory, which is one of the main goals of this paper.
  
\begin{figure}
 \centering 
  \scalebox{.40}{\includegraphics[width=\textwidth,scale=0.2]{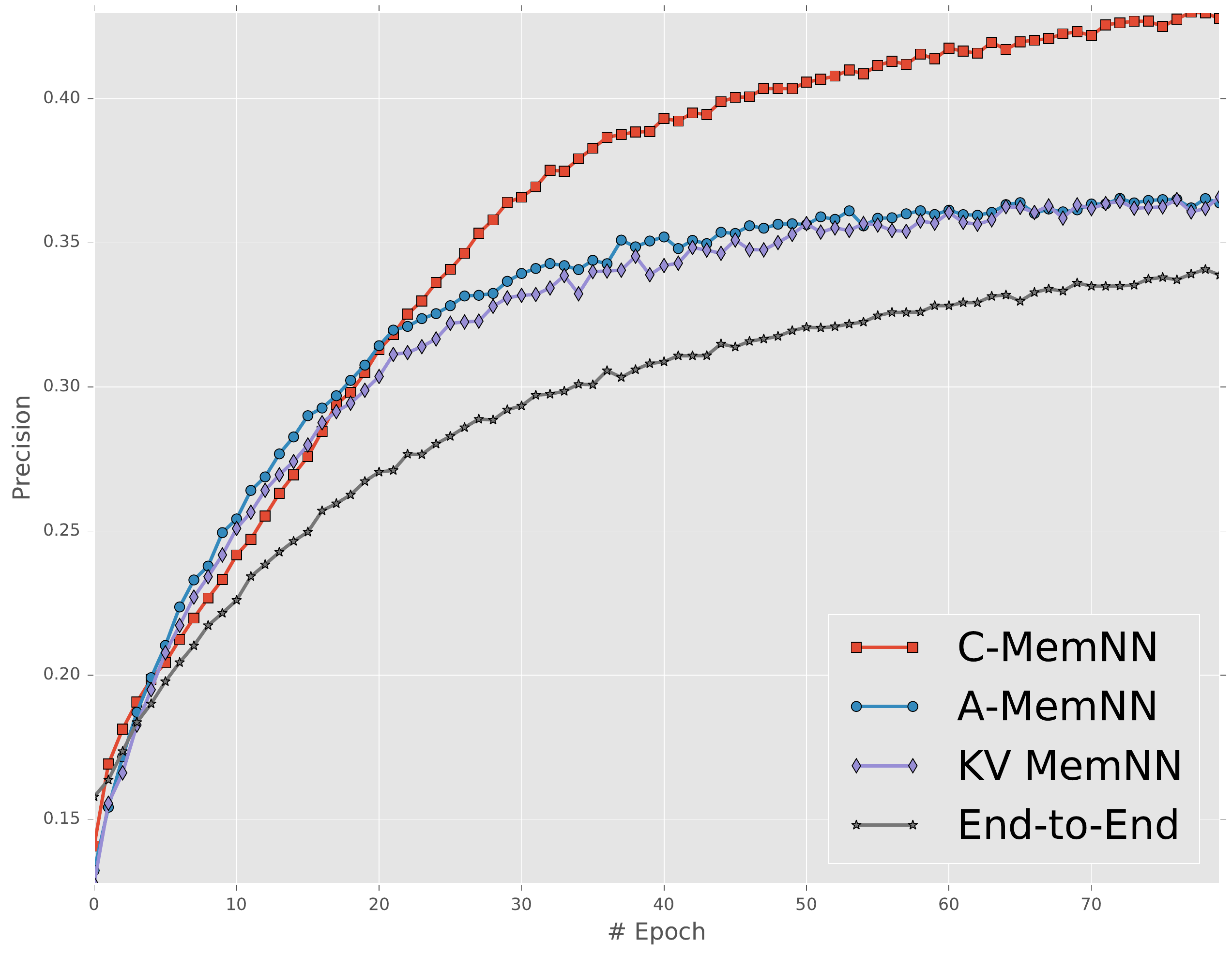} }
  \caption{Precision@5 plot for various models on validation data (at 4 hops).}
\end{figure}

\begin{table*}[tbp]
\centering

\label{my-label}
\def\arraystretch{1.20}
\begin{tabular}{clcccccccc}
 &  & \multicolumn{3}{c}{\# classes = 50} &   & & \multicolumn{3}{c}{\# classes = 100}     \\
  \bottomrule

\# \texttt{Hops} & \multicolumn{1}{c}{\texttt{Model}} & \multicolumn{1}{c}{\begin{tabular}[c]{@{}c@{}}\texttt{AUC}  \\ \texttt{(macro)} \\ $\uparrow$\end{tabular}} & \multicolumn{1}{c}{\begin{tabular}[c]{@{}c@{}}\texttt{Average} \\ \texttt{Precision}\\ @5 $\uparrow$ \end{tabular}} & \multicolumn{1}{c}{\begin{tabular}[c]{@{}c@{}}\texttt{Hamming} \\ \texttt{Loss} \\ $\downarrow$ \end{tabular}} & \multicolumn{1}{c|}{} &  & \multicolumn{1}{c}{\begin{tabular}[c]{@{}c@{}}\texttt{AUC} \\ \texttt{(macro)}\\ $\uparrow$ \end{tabular}} & \multicolumn{1}{c}{\begin{tabular}[c]{@{}c@{}}\texttt{Average}\\ \texttt{Precision}\\ @5 $\uparrow$\end{tabular}} & \multicolumn{1}{c}{\begin{tabular}[c]{@{}c@{}}\texttt{Hamming}\\ \texttt{Loss} \\ $\downarrow$ \end{tabular}} \\
 
 \bottomrule
 
\multirow{4}{*}{3} & End-to-End & 0.759 & 0.32 & 0.06 & \multicolumn{1}{l|}{} &  & 0.664 & 0.23 & 0.15 \\
 & KV MemNN & 0.761 & 0.36 & \textbf{0.05} & \multicolumn{1}{l|}{} &  & 0.679 & 0.24 & 0.14 \\
 & A-MemNN & 0.762 & 0.36 & 0.06 & \multicolumn{1}{l|}{} &  & 0.675 & 0.23 & 0.14 \\
 & C-MemNN & \textbf{0.785} & \textbf{0.39} & \textbf{0.05} & \multicolumn{1}{l|}{} &  & \textbf{0.697} & \textbf{0.27} & \textbf{0.12} \\
  \bottomrule
\multirow{4}{*}{4} & End-to-End & 0.760 & 0.33 & 0.04 & \multicolumn{1}{l|}{} &  & 0.672 & 0.24 & 0.15 \\
 & KV MemNN & 0.776 & 0.35 & 0.04 & \multicolumn{1}{l|}{} &  & 0.683 & 0.24 & 0.13 \\
 & A-MemNN & 0.775 & 0.37 & 0.03 & \multicolumn{1}{l|}{} &  & 0.689 & 0.23 & 0.11 \\
 & C-MemNN & \textbf{0.795} & \textbf{0.42} & \textbf{0.02} & \multicolumn{1}{l|}{} &  & \textbf{0.705} & \textbf{0.27} & \textbf{0.09} \\
  \bottomrule
\multirow{4}{*}{5} & End-to-End & 0.761 & 0.34 & 0.04 & \multicolumn{1}{l|}{} &  & 0.683 & 0.25 & 0.14 \\
 & KV MemNN & 0.775 & 0.36 & 0.03 & \multicolumn{1}{l|}{} &  & 0.697 & 0.25 & 0.11 \\
 & A-MemNN & 0.804 & 0.40 & 0.02 & \multicolumn{1}{l|}{} &  & 0.720 & 0.29 & 0.11 \\
 & C-MemNN & \textbf{0.833} & \textbf{0.42} & \textbf{0.01} & \multicolumn{1}{l|}{} &  & \textbf{0.767} & \textbf{0.32} & \textbf{0.05} \\
 \bottomrule
\end{tabular}

\caption{Evaluation results of various memory networks on \texttt{MIMIC-III} dataset.}
\end{table*}

\subsection{Training}
    We use Adam~\cite{Kingma2014AdamAM} stochastic gradient descent for optimizing the learned parameters. The learning rate is set to 0.001 and batch size for each iteration to 100 for all models. For the final prediction layer, we use a fully connected layer on top of the output from equation~\ref{eq:probs} with a sigmoid activation function. The loss function is the sum of cross entropy from prediction labels and prediction memory slots using addressing schema. Complexity of the model was penalized by adding $L2$ regularization to the cross entropy loss function.  We use dropout~\cite{Srivastava2014DropoutAS} with probability 0.5 on the output-to-decision sigmoid layer and limit the norm of the gradients to be below 20. Models are trained on $80\%$ of the data and validated on $10\%$. The remaining $10\%$ is used as test set which is evaluated only once across all experiments with different models.

\subsection{Results and Analysis}

We present experiments in which performance is evaluated using three metrics: the area under the ROC curve (AUC), the average precision over the top five predictions, and the hamming loss. The AUC is calculated by taking unweighted mean of the AUC values for each label - this is also known as the macro AUC\@. Average precision over the top five predictions is reported because it is a relevant metric for real world applications. Hamming loss is reported instead of accuracy because it is a better measure for multi-label classification~\cite{elisseeff2001kernel}. 

    As shown in Table 3, C-MemNN is able to exceed the results of various other memory networks across all experiments.  The improvement is more pronounced with a higher number of memory hops. This is because of the learning saturation of vanilla memory networks over multiple hops. While A-MemNN has better results for higher hops it does not improve upon KV-MemNN at lower hops. The strength of our model lies at higher hops, as the condensed memory state $\widetilde{u}$ after several hops contains more information than the same size input memory state $u$. Across all models, results improve as the number of hops increases, although with diminishing returns. The AUC value of C-MemNN with five memory hops for 100 labels is higher than the AUC value for End-to-End models trained only for three hops, which shows efficient training of higher hops produces good results. 
    
    Most documents do not have five labels (Figure 1) and thus precision obtained for five predictions is poor across all models. Hamming Loss correlates very well with other metrics along with the cross-entropy loss function, which was used for training. 

    Our model has 30\% more parameters than standard Memory Networks ($\boldsymbol{D}$ for every pair of $\boldsymbol{A}$ and $\boldsymbol{C}$). Figure 3 shows that adding more memory does not lead to overfitting. Training time of our model for GPU implementation is same as other memory networks, as condensed memory state ($\widetilde{u}$) is trained almost in parallel with standard memory state ($u$) for every hop. 

\section{Conclusion and Future Work}
\citeauthor{DBLP:journals/corr/WestonCB14} (2014) discussed the possibility of a better memory representation for complex inferencing tasks. We achieved a better memory representation by condensing the previous hops in a novel way to obtain a hierarchical representation of the internal memory. We have shown the efficacy of the proposed memory representation for clinical diagnostic inferencing from raw textual data. We discussed the limitations of memory networks for multi-label classification and explored gated addressing to achieve a better mapping between the clinical notes and the memory slots. We have shown that training multiple hops with condensed representation is helpful, but this is still computationally expensive. We plan to investigate asynchronous memory updating, which will allow for faster training of memory networks. In the future, we will explore other knowledge sources and the recently proposed word vectors for the biomedical domain~\cite{chiu2016train}.

\section*{Acknowledgments}
The authors would like to thank the anonymous reviewers for their valuable comments and feedback. The first author is especially grateful to Prof. James Storer, Brandeis University, for his guidance, and Nick Moran, Ryan Marcus and Solomon Garber for helpful discussions.

\bibliographystyle{aaai}
\bibliography{refs}
\end{document}